\pdfoutput=1

\documentclass[11pt]{article}

\usepackage{EMNLP2022}

\usepackage{times}
\usepackage{latexsym}

\usepackage[T1]{fontenc}

\usepackage[utf8]{inputenc}

\usepackage{microtype}

\usepackage{inconsolata}

\usepackage{graphicx}

\usepackage{enumitem}
\usepackage{comment}
\usepackage{multirow}
\usepackage{makecell}
\usepackage{hyperref}

\usepackage{soul}

\newcommand\ie{{\it i.e.}}
\newcommand\eg{{\it e.g.}}

\newcommand\dg{{\sc dg}}
\newcommand\ood{\textsc{ood}}
\newcommand\erm{\textsc{erm}}
\newcommand\nlp{{\sc nlp}}
\newcommand\qa{{\sc qa}}
\newcommand\kd{\textsc{kd}}
\newcommand\rc{{\sc rc}}

\newcommand\squad{\textsc{sq}u\textsc{ad}}
\newcommand\newsqa{\textsc{n}ews\textsc{qa}}
\newcommand\nq{\textsc{nq}}
\newcommand\hotpotqa{\textsc{h}otpot\textsc{qa}}
\newcommand\triviaqa{\textsc{t}rivia\textsc{qa}}
\newcommand\searchqa{\textsc{s}earch\textsc{qa}}

\newcommand\bioasq{\textsc{b}io\textsc{asq}}
\newcommand\drop{\textsc{drop}}
\newcommand\duorc{\textsc{d}uo\textsc{rc}}
\newcommand\race{\textsc{race}}
\newcommand\relex{\textsc{r}el\textsc{e}x}
\newcommand\textbookqa{\textsc{t}extbook\textsc{qa}}
\newcommand\mldg{\textsc{mldg}}
\newcommand\bert{\textsc{bert}}
\newcommand\dil{\textsc{dil}}

\title{\textit{Not to Overfit or Underfit the Source Domains?}
\\ An Empirical Study of Domain Generalization in Question Answering}

\author{Md Arafat Sultan \quad Avirup Sil \quad Radu Florian\\ IBM Research AI \\ arafat.sultan@ibm.com,
\{avi, raduf\}@us.ibm.com}

\begin{document}
\maketitle
\begin{abstract}
Machine learning models are prone to overfitting their training (source) domains, which is commonly 
believed to be 
the reason 
why 
they falter 
in novel target domains.
Here we examine the contrasting view that multi-source domain generalization (\dg{}) is first and foremost a problem of mitigating source domain \emph{underfitting:} models not adequately learning the signal already present in their multi-domain training data.
Experiments on a reading comprehension \dg{} benchmark show that as a model
learns its source domains better---using familiar methods such as knowledge distillation (\kd{}) from a bigger model---its zero-shot out-of-domain utility improves at an even faster pace.
Improved source domain learning also demonstrates superior out-of-domain generalization over three popular existing \dg{} 
approaches that aim to limit overfitting.
Our implementation of \kd{}-based domain generalization is available via PrimeQA at: \url{https://ibm.biz/domain-generalization-with-kd}.
\end{abstract}

\section{Introduction}
\label{section:introduction}
Domain generalization (\dg{}) seeks to train models on a small number of \emph{source} domains 
in a way that maximizes their zero-shot out-of-domain (\ood{}) utility  \cite{blanchard2011generalizing,muandet2011domain}.
Many existing \dg{} methods are rooted in the premise that weak generalization under domain shift
occurs due to 
``source domain overfitting'': the learning of spurious source domain correlations, \ie{}, noise, that are unrelated to the core learning task, and are therefore unlikely to be present in novel target domains.
The popular \textit{domain-invariant learning} (\dil{}) paradigm, for example, proposes to limit overfitting by imposing some form of cross-domain regularization on the  empirical risk minimization (\erm{}) process over labeled multi-domain training data \cite{wang2021generalizing,zhou2021domain}.


We argue in this paper that this emphasis on \textit{not fitting the noise} 
minimizes the importance of \textit{actually learning the signal} present in the source domains, \ie{}, \textbf{\textit{not underfitting}} them, for \dg{}.
Underfitting occurs when a trained model, due to 
inadequacies in its capacity or the training procedure, fails to learn training set patterns that are truly representative of the task.
In a recent study of \dg{} in computer vision, 
\citet{gulrajani2021search} find that 
large models with sufficiently high capacity can demonstrate strong \ood{} generalization even with ordinary \erm{} training---without the need for \dil{}---when properly configured.
Here we ask a very similar question, but (\textit{a}) in the \nlp{} context of question answering (\qa{}), and (\textit{b}) for relatively small models that are more prone to underfitting, \eg{}, a \bert{}-\textit{base} \qa{} model \cite{devlin2019bert}.
Concretely, we investigate if enabling smaller \qa{} models to learn their source domains better also improves their generalization to new domains and how the results compare to \dil{}.

This formulation essentially views \ood{} generalization in low-capacity models as a natural extension of the more familiar and arguably simpler problem of in-domain generalization in a multi-domain setting.
A key advantage of such an approach is that familiar supervised learning methods such as knowledge distillation (\kd{}) \cite{hinton2014distilling}---which was specifically designed for the very purpose of training small yet high-accuracy models---can now be leveraged for \dg{}.
\kd{} generally provides stronger source domain supervision than \erm{} by minimizing a surrogate risk, which utilizes the soft predictions of a larger \textit{teacher} model (\eg{}, a \bert{}-\textit{large} \qa{} model) as the learning targets for the smaller model being trained, now called the \textit{student}.
Additionally, synthesized input in large quantities has been found to further enhance the performance of \kd{}
 \cite{chen2020improved,liang2021mixkd}.
Here we extend the application of these methods to \dg{} for \qa{}.

We evaluate our methods on a multi-dataset reading comprehension benchmark \cite{fisch2019mrqa} and compare their accuracy with three popular \dil{} approaches: domain adversarial training \cite{ganin2016domain,lee2019domain}, episodic training \cite{li2019domain}---for which we propose a novel variant suitable for deep transformer models---and meta learning \cite{finn2017model,li2018learning}.
We also design experiments to answer more targeted questions such as: (1)~Are the improvements more prominent on in-domain validation data than on \ood{} test instances, which could be indicative of weak generalization? (2)~Do the proposed methods falter on input cases where domain-invariant approaches thrive, potentially indicating weakness on extremely distant test cases?
In all these different evaluations, our methods exhibit far superior \dg{} than the three existing methods, whereas the latter only marginally outperform \erm{}.

While more experiments on additional tasks, datasets and baselines are needed before a firm conclusion can be reached on the superiority of the proposed formulation of \dg{} over \dil{} (or vice versa), our findings do point to a need for a better understanding of optimal source domain learning as an approach to \dg{}.
A primary goal of this paper is to motivate future explorations of this important research question.

\section{Methodology}
\label{section:methods}
This section describes the proposed \kd{}-based \dg{} approaches as we apply them to the problem of reading comprehension.

\subsection{The Reading Comprehension Task}
Given a question $q$ and a passage $p$ that answers $q$, reading comprehension (\rc{}) outputs the answer $a$.
In \textit{extractive} \rc{}, which is the form used in our experiments, $a$ is assumed to be a subtext of $p$; the goal is therefore to locate $a$ in $p$.

\subsection{Multi-Dataset Knowledge Distillation}
\label{subsection:kd}
For improved multi-domain training, we rely on knowledge distillation (\kd{}), which naturally trains small yet highly accurate models by leveraging the predictions of a larger teacher model.
We first train a single multi-domain teacher using \erm{} on labeled data from all source domains.
We follow standard \rc{} training procedure for this step, which separately trains an answer start and an end predictor, as described in \cite{devlin2019bert}.

The knowledge of this teacher is then distilled into a smaller student model---equal in size with the baselines---by minimizing the following \textsc{mse} loss on the same set of training examples:
\begin{equation}
    \mathcal{L}_{KD} = \|z_s(x) - z_t(x) / \mathcal{T}\|_2^2
    \label{equation:kd}
\end{equation}
where $z_s(x)$ and $z_t(x)$ are the logits computed for an input question-passage pair $x$ by the student and the teacher, respectively; $\mathcal{T}$ is the temperature.
Two separate \kd{} losses are again minimized per training example, one for the start and the other for the end of the answer.

\subsection{Augmenting \kd{} with Synthetic Questions}
To facilitate the distillation of further knowledge from the teacher, we synthesize additional questions using a sequence-to-sequence model.
An encoder-decoder language model \cite{lewis2020bart,raffel2020exploring} is first fine-tuned for each source domain, where question-passage pairs from the corresponding dataset constitute the training examples: the passage is the source and the question is the target.
Note that we use the teacher's soft answer predictions as targets during \kd{} (Eq.~\ref{equation:kd}), and therefore do not need to provide any answers as part of the synthetic data.

\citet{sultan2020importance} demonstrate that sampling-based generation, \eg{}, with a top-$p$ top-$k$ sampler, produces more useful synthetic training data than deterministic approaches like greedy or beam search decoding.
Moreover, \citet{chen2020improved} find that (\textit{a}) large amounts of diverse synthetic data can be more effective at supporting \kd{} than typically smaller amounts of human-labeled \textit{gold} examples, and (\textit{b}) leveraging both in a two-stage process yields the best results.
We incorporate these suggestions into our work by sampling examples from our generators and performing \kd{} with first synthetic and then gold training data.
Unlike those earlier studies, however, we apply the above procedure to the multi-source \dg{} problem.


\section{Experiments}
\label{section:experiments}
Here we describe the baseline \dil{} approaches our proposed methods are evaluated against, our experimental setup and results.

\subsection{Baselines}
\label{subsection:dil}
We select three existing \dil{} methods as baselines on account of their recency, popularity and general applicability to different machine learning problems. 
Each of these methods imposes additional requirements on top of \erm{} to incorporate domain invariance into the trained models.

For \textbf{\textit{domain-adversarial training}} \cite{ganin2016domain}, that added requirement is for the model to learn domain-agnostic hidden representations of the training inputs.
This is accomplished by training a domain classifier in parallel with the \rc{} model and teaching their shared feature extractor to produce adversarial representations for the domain classifier \cite{lee2019domain}.

Given a model with parameters $\Theta$ to be optimized on mult-domain training data, \textbf{\textit{episodic training}} \cite{li2019domain} updates a random subset $\Theta{'}$ in each iteration; values for the remaining parameters $\Theta\setminus\Theta{'}$ are copied over from a weaker model pre-trained on one of the source domains other than that of the current example.
This procedure effectively forces the $\Theta{'}$ subnetwork to become more robust to domain shift as it learns to work with an \ood{} companion $\Theta\setminus\Theta{'}$.
While \citet{li2019domain} use a fixed breakdown of $\Theta$ into a feature extractor and a task head, we relax this condition for multilayer transformer models to allow a split after a randomly chosen transformer layer.

Finally, \textbf{\textit{meta learning}} \cite{finn2017model} for \dg{} (\mldg{}) \cite{li2018learning} uses disjoint subsets of the source domains as {\it meta-train} and {\it meta-test} domains at each training step.
It uses the meta-train set to update model parameters in a way that improves performance on the meta-test set.
This is accomplished using a second-order differentiation through the parameter updates of the model.

\subsection{Setup}
We run our experiments on the the public subset of the \dg{} benchmark by \citet{fisch2019mrqa}\footnote{\scriptsize\url{https://github.com/mrqa/MRQA-Shared-Task-2019\#datasets}}.
It consists of ({\it a})~training and in-domain validation data from six source datasets, and ({\it b})~six target datasets for evaluation.
Table~\ref{table:datasets} shows some key statistics.
We refer the reader to the original paper for a detailed description of each dataset.

Given our stated intent to study \dg{} with relatively small models (\S\ref{section:introduction}), we fine-tune \bert{}-\textit{base} (110\textsc{m} parameters) \cite{devlin2019bert} for \rc{} with the different training methods \cite{wolf2020transformers}.
In \kd{} experiments, we use a \bert{}-\textit{large} teacher (345\textsc{m} parameters) fine-tuned using \erm{} on the source datasets.
To prevent any confounding effects from negative windows \cite{fisch2019mrqa}, only those sliding windows of the training contexts are retained that actually contain an answer.
We upsample from smaller training sets to make the number of examples equal across domains, and include examples from only one training set in each mini-batch (which worked better than multi-dataset mini-batches in our experiments).
All hyperparameters are tuned by optimizing the macro-averaged F1 score on the in-domain validation sets.
See Appendix~\ref{appendix:model-selection} for more details.

\begin{table}[t]
\small
\centering
\begin{tabular}{l|cc}
\multicolumn{1}{c|}{\textbf{Source}} & \textbf{Train} & \textbf{Dev} \\
\Xhline{2\arrayrulewidth}
\squad{}~\cite{rajpurkar2016squad} & 86,588 & 10,507 \\
\newsqa{}~\cite{trischler2017newsqa} & 74,160 & 4,212 \\
\nq{}~\cite{kwiatkowski2019natural} & 104,071 & 12,836 \\
\hotpotqa{}~\cite{yang2018hotpotqa} & 72,928 & 5,904 \\
\triviaqa{}~\cite{joshi2017triviaqa} & 61,688 & 7,785 \\
\searchqa{}~\cite{dunn2017searchqa} & 117,384 & 16,980 \\
\multicolumn{3}{c}{} \\
\end{tabular}
\begin{tabular}{l|c}
\multicolumn{1}{c|}{\textbf{Target}} & \textbf{Test} \\
\Xhline{2\arrayrulewidth}
\bioasq{}~\cite{tsatsaronis2015overview} & 1,504 \\
\drop{}~\cite{dua2019drop} & 1,503 \\
\duorc{}~\cite{saha2018duorc} & 1,501 \\
\race{}~\cite{lai2017race} & 674 \\
\relex{}~\cite{levy2017zero} & 2,948 \\
\textbookqa{}~\cite{kembhavi2017you} & 1,503 \\
\end{tabular}
\caption{Dataset statistics{\scriptsize ~\cite{fisch2019mrqa}}: $\#$ of examples.}
\label{table:datasets}
\end{table}

With every training method, we train six models, each on a unique five-set combination of the six source datasets.
Each of these six models is selected using the validation sets of the same five datasets.
The performance of the method on an individual test set is the mean F1 score of these six models on the set.
Finally, the F1 scores on all six test sets are macro-averaged to measure the method's overall \ood{} performance.

For synthetic data generation, we fine-tune separate \textsc{bart}-\textit{large} models \cite{lewis2020bart} on the individual source datasets.
We generate 500k questions per dataset from Wikipedia contexts using \mbox{top-$p$} \mbox{top-$k$} sampling ($p$=.95, $k$=10).

\begin{table*}[ht]
\small
\centering
\begin{tabular}{l|cccccc|c}
\multicolumn{1}{c|}{\multirow{2}{*}{\textbf{Method}}} & \multicolumn{6}{c|}{\textbf{Test Set}} & \multirow{2}{*}{\textbf{Avg.}} \\ 
\cline{2-7}
& \bioasq{} & \drop{} & \duorc{} & \race{} & \relex{} & \textbookqa{} & \\
\Xhline{2\arrayrulewidth}
\textsc{erm} & 51.7\scriptsize{$\pm$1.4} & 37.8\scriptsize{$\pm$0.2} & 55.1\scriptsize{$\pm$1.5} & 39.0\scriptsize{$\pm$1.0} & 83.3\scriptsize{$\pm$0.3} & 51.6\scriptsize{$\pm$1.8} & 53.1\scriptsize{$\pm$1.0} \\
\hline
\multicolumn{8}{l}{\textit{Domain-invariant learning to \textbf{limit overfitting}:}} \\
\hline
Domain-Adv & 51.7\scriptsize{$\pm$1.9} & 38.4\scriptsize{$\pm$1.1} & 56.0\scriptsize{$\pm$1.0} & 39.3\scriptsize{$\pm$0.6} & 83.5\scriptsize{$\pm$0.7} & 52.2\scriptsize{$\pm$1.7} & 53.5\scriptsize{$\pm$1.2} \\
Episodic & 52.0\scriptsize{$\pm$1.6} & 38.4\scriptsize{$\pm$1.4} & 56.4\scriptsize{$\pm$1.1} & 40.1\scriptsize{$\pm$0.5} & 83.3\scriptsize{$\pm$0.4} & 52.3\scriptsize{$\pm$1.9} & 53.7\scriptsize{$\pm$1.2} \\
\mldg{} & 52.7\scriptsize{$\pm$0.9} & 38.0\scriptsize{$\pm$1.1} & 56.1\scriptsize{$\pm$1.5} & 39.5\scriptsize{$\pm$0.9} & 83.9\scriptsize{$\pm$0.3} & 51.1\scriptsize{$\pm$2.0} & 53.6\scriptsize{$\pm$1.1} \\
\hline
\multicolumn{8}{l}{\textit{Improved source domain learning to \textbf{address underfitting}:}} \\
\hline
\kd{} (gold-only) & 53.2\scriptsize{$\pm$0.7} & 42.2\scriptsize{$\pm$1.3} & 58.4\scriptsize{$\pm$1.6} & 42.9\scriptsize{$\pm$0.9} & 84.1\scriptsize{$\pm$0.9} & 56.1\scriptsize{$\pm$2.6} & 56.2\scriptsize{$\pm$1.3} \\
\kd{} (augmented) & \textbf{53.4}\scriptsize{$\pm$0.9} & \textbf{45.2}\scriptsize{$\pm$1.7} & \textbf{60.3}\scriptsize{$\pm$1.2} & \textbf{44.2}\scriptsize{$\pm$0.8} & \textbf{84.8}\scriptsize{$\pm$0.6} & \textbf{58.0}\scriptsize{$\pm$1.4} & \textbf{57.6}\scriptsize{$\pm$1.1} \\
\end{tabular}
\caption{
Performance (F1 score) of different training methods on \ood{} test data.
Each score is a mean$\pm$\textsc{sd} over six models, each trained on a unique five-set combination of the six source datasets.
While the domain-invariant methods provide small gains over plain \erm{}, improved source domain learning demonstrates by far the best results.
}\label{table:main-results}
\end{table*}

\subsection{Results}
In Table~\ref{table:id_vs_ood} (column 1), we show in-domain validation results for \erm{} and the two variants of \kd{}: \erm{} clearly exhibits some underfitting relative to the other two methods, as it has the lowest score of the three.
\kd{} using gold instances helps mitigate the underfitting effect to some extent; augmented \kd{} with additional synthetic questions further improves results.
These results confirm the utility of each of our methods in improving in-domain generalization of \rc{} systems.

Table~\ref{table:main-results} summarizes the performance of different training methods on the \ood{} test sets.
The \dil{} methods do yield small gains over \erm{}, but improved source domain learning with \kd{} and data augmentation outperforms by a much larger margin.
Crucially, we observe improvements on each individual test set.
In a two-tailed Student's $t$-test, we find all differences between (\textit{a}) \dil{} and \erm{}, and (\textit{b}) \kd{}-based methods and \dil{}, to be statistically significant ($p<.0001$).  

\begin{table}[t]
\small
\centering
\begin{tabular}{l|cc}
\multicolumn{1}{c|}{\textbf{Method}} & \textbf{\textsc{id}-Dev} & \textbf{\ood{}-Test} \\
\Xhline{2\arrayrulewidth}
\erm{} & 75.0 & 53.1 \\
\hline
\kd{} (gold-only) & 76.4 \mbox{\scriptsize (1.9\%)} & 56.2 \mbox{\scriptsize (\textbf{5.8\%})} \\
\kd{} (augmented) & 77.2 \mbox{\scriptsize (2.9\%)} & 57.6 \mbox{\scriptsize (\textbf{8.5\%})} \\
\end{tabular}
\caption{F1 score (and relative gain over \erm{}) for each proposed \kd{}-based method. Gains on the \ood{} test sets outpace those on the in-domain dev sets, indicating strong generalization.}
\label{table:id_vs_ood}
\end{table}

To further examine the generalization induced by \kd{}, we compare in Table~\ref{table:id_vs_ood} their performance on in-domain validation data versus \ood{} test data: 
both methods provide substantially larger relative gains (5.8--8.5$\%$) over \erm{} in \ood{} evaluation than in in-domain evaluation (1.9--2.9$\%$).
These results clearly indicate that many of the new and/or improved patterns the proposed methods teach do indeed generalize to other domains.

\begin{table}[h]
\small
\centering
\begin{tabular}{l|c}
\multicolumn{1}{c|}{\textbf{Method}} & \textbf{Avg. F1} \\
\Xhline{2\arrayrulewidth}
\kd{} & \textbf{56.2} \\
\hline
\kd{} + Domain-Adv & 55.4 \\
\kd{} + Episodic & 55.5 \\
\kd{} + \mldg{} & 55.8 \\
\end{tabular}
\caption{Domain-invariant learning does not complement \kd{}-based source domain learning in \ood{} tests.}
\label{table:erm_oodg_complementarity}
\end{table}

Even though the \kd{}-based methods exhibit stronger overall \ood{} generalization, it is still possible that \dil{} teaches certain \dg{}-inducing patterns that even powerful source domain supervision fails to reveal, in which case the former should complement the latter well.
To test this effect, we train three models using each of the three \dil{} methods, but replace \erm{} with \kd{} (gold instances only) as the underlying training mechanism for \rc{}.
As Table~\ref{table:erm_oodg_complementarity} shows, none of the three combinations does better than \kd{} alone. 
Along with the results of Table~\ref{table:main-results}, this result indicates that as a learner is exposed to more powerful source domain supervision, \dil{} starts to lose its ability to complement its already strong \ood{} generalization.

\begin{figure}[t]
\centering
\includegraphics[width=1\linewidth]{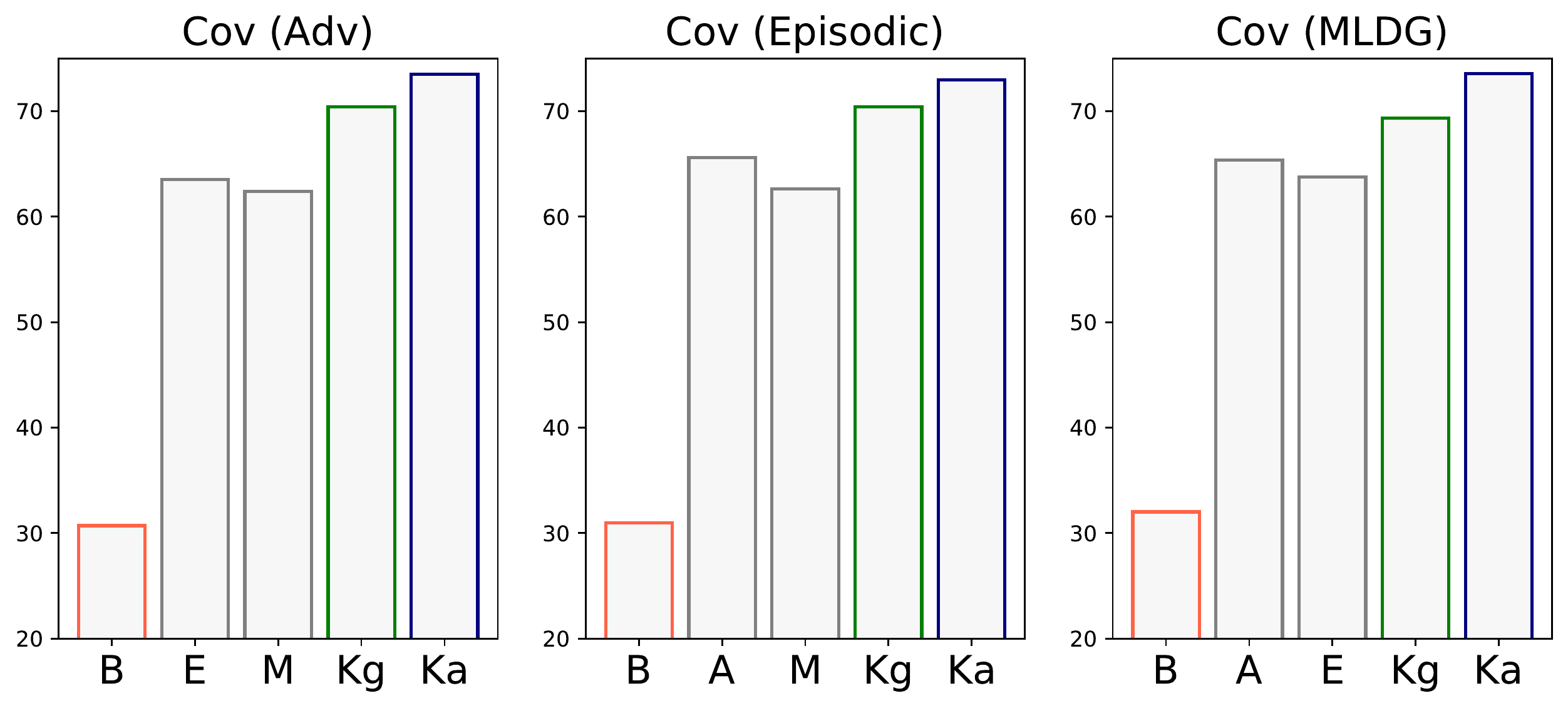}
\caption{The ability of different methods to \textit{cover}, \ie{}, function as a proxy for, other methods. The leftmost plot shows the coverage of domain-adversarial training by other methods, for example.
\kd{}-based source domain learning provides the best coverage across the board. B:~\erm{}, E:~Episodic, M:~\mldg{}, A:~Domain-Adv, Kg:~gold-only \kd{}, Ka:~\kd{} combined with synthetic data augmentation.}
\label{figure:coverage}
\end{figure}

As a final test of adequacy for multi-source \kd{} as a \dg{} method, we look at its ability to function as a proxy for the different \dil{} methods.
Let $\mathcal{E}$ be the set of examples for which a training method \textsc{m} has a higher F1 score than  \erm{}, representing the \dg{} capabilities of \textsc{m}.
We define the \textit{coverage} of $\textsc{m}$ by another method $\textsc{m}{'}$ as the relative F1 score of $\textsc{m}{'}$ as a fraction ($\%$) of the F1 score of \textsc{m} on $\mathcal{E}$.
This metric essentially quantifies the degree to which the \dg{} capabilities of \textsc{m} is retained by $\textsc{m}{'}$.
As the bar charts of Figure~\ref{figure:coverage} demonstrate, the \kd{}-based methods provide the best coverage of all three \dil{} methods; the latter, while providing considerably better coverage of one another than \erm{}, lag behind \kd{} in all three cases.
These results again suggest that strong source domain learning may potentially be a sufficiently optimal policy for multi-source \dg{}, without the need for an explicit enforcement of domain invariance.

\section{Conclusion}
\label{section:conclusion}
This paper puts forward the view with empirical evidence for question answering (\qa{}) that contrary to popular belief, multi-source domain generalization (\dg{}) is better modeled as a problem of addressing model underfitting than overfitting.
Our experimental results show that by simply learning the training domains well via knowledge distillation, even when the number of such domains is relatively small, strong out-of-domain (\ood{}) generalization can be achieved in \qa{} without the need for any cross-domain regularization.
These findings point to the need for a re-examination of whether \textit{not fitting the noise} in training data is indeed a more reasonable path forward for \dg{} than actually \textit{learning the signal} in it; we hope that our work will inspire future efforts to answer this important question in greater detail with a wider range of tasks and approaches.

\section*{Limitations}
We explore the problem of multi-source domain generalization (\dg{}) in \qa{} with new and existing methods.
We believe that our findings will generalize to more baselines and datasets, but here we only show proof of concept for a select set of existing baselines and a single multi-dataset \dg{} benchmark.

\bibliography{custom}
\bibliographystyle{acl_natbib}

\clearpage
\appendix
\section{Appendix}
\label{appendix}

\subsection{Qualtitative Analysis}
\label{appendix:qualitative}
To better understand what new patterns the proposed method of augmented knowledge distillation (\kd{}-Aug) teaches, we take a closer look at a random sample of the test instances where the baseline model has an F1 score of zero and the student has an F1 score greater than .5.
Table~\ref{table:qualitative} shows four such examples from four different test sets.
Even in this very small sample, we observe a number of different ways in which the \kd{}-Aug student is better than the plain \erm{} baseline:

\begin{itemize}
\item \bioasq{}: The student has learned that synonyms are commonly mentioned in parentheses, especially in the beginning of a sentence.
\item \duorc{}: The word ``overdose'' can be difficult to model in context as the subject of the verb ``overdose'' becomes the object in the expression ``giving someone an overdose''. The student appears to have learned this nuanced semantics, which the baseline model has not.
\item \textsc{r}elation\textsc{e}xtraction: Unlike the baseline, the student knows that a fictional universe is not a time period, and the word ``Universe'' at the end of the phrase ``DC Universe'' is a strong hint to capitalize on in this particular case. Crucially, it has also learned to ignore the spurious surface-level match in this example between the question and the phrase ``from the 30th century'' in the passage.
\item \drop{}: This one is a somewhat harder case to analyze, where a plausible explanation could be that the student has better knowledge about people's names and only the last name ``Gostkowski'' was sufficient for it to recognize it as a player's name.
\end{itemize}

Even though these patterns were taught by \kd{}-Aug using a small number of source domains, their domain-agnostic importance is quite clear and intuitive, which is also supported by the experiments reported in this paper. 

\begin{table*}
\centering
\small
\begin{tabular}{p{14cm}}
\Xhline{2\arrayrulewidth}
\textcolor{darkblue}{\textbf{Dataset}}: \bioasq{} \\
\textcolor{darkblue}{\textbf{Question}}: Name synonym of Acrokeratosis paraneoplastica.\\
\textcolor{darkblue}{\textbf{Passage}}: Acrokeratosis paraneoplastica ( \hl{Bazex ' syndrome} ) is a rare but clinically distinctive dermatosis that has been associated in all reported cases , to our knowledge , with either a primary malignant neoplasm of the upper aerodigestive tract or metastatic cancer to the lymph nodes of the neck . Acrokeratosis paraneoplastica was found in a 53-year - old black man with squamous cell carcinoma of the tonsil . A distinctive series of changes was found on histopathologic examination of biopsy specimens taken from his skin lesions , and direct immunofluorescence microscopy of both lesional and nonlesional skin specimens showed immunoglobulin and complement deposition on the epidermal basement membrane . The skin lesions largely resolved following radiation therapy of the neoplasm and of the presumably involved lymph nodes . \\
\textcolor{darkblue}{\textbf{\textsc{gt}}}: [`Bazex syndrome'] \\
\textcolor{darkblue}{\textbf{\erm{} Answer:}}  dermatosis \\
\textcolor{darkblue}{\textbf{\kd{}-Aug Answer:}}  Bazex ' syndrome \\
\hline
\textcolor{darkblue}{\textbf{Dataset}}: \duorc{} \\
\textcolor{darkblue}{\textbf{Question}}: Who overdoses on insulin? \\
\textcolor{darkblue}{\textbf{Passage:}} The film tells the story of a psychiatrist , Dr. Cross ( Vincent Price ) , who is treating a young woman , \hl{Janet Stewart} ( Anabel Shaw ) , who is in a coma - state , brought on when she heard loud arguing , went to her window and saw a man strike his wife with a candlestick and kill her . It also stars Lynn Bari as Dr. Cross 's nurse / lover , Elaine Jordan . As Stewart comes out of her shock , she recognizes Dr. Cross as the killer . He then takes her to his sanitarium and at Elaine 's urging , gives \hl{Janet} an overdose of insulin under the pretense of administering insulin shock therapy . He ca n't bring himself to murder her in cold blood , though , and asks Elaine to get the medicine to save her . Elaine refuses , they argue , and he strangles her . A colleague of Dr. Cross , Dr. Harvey , saves Janet 's life and Dr. Cross is taken into custody by a lawyer from the District Attorney 's office . \\
\textcolor{darkblue}{\textbf{\textsc{gt:}}} [`Janet.', `Janet'] \\
\textcolor{darkblue}{\textbf{\erm{} Answer:}} Dr. Cross \\
\textcolor{darkblue}{\textbf{\kd{}-Aug Answer:}} Janet Stewart \\
\hline
\textcolor{darkblue}{\textbf{Dataset}}: RelationExtraction \\
\textcolor{darkblue}{\textbf{Question}}: What is the name of the fictional universe that Polar Boy is from? \\
\textcolor{darkblue}{\textbf{Passage:}} Polar Boy is a fictional character from the 30th century of the \hl{DC Universe} , initially suggested by reader Buddy Lavigne of Northbrook , Illinois in the letters page of Adventure Comics \# 304 , January , 1963 . \\
\textcolor{darkblue}{\textbf{\textsc{gt:}}} [`DC Universe'] \\
\textcolor{darkblue}{\textbf{\erm{} Answer:}} 30th century \\
\textcolor{darkblue}{\textbf{\kd{}-Aug Answer:}} DC Universe \\
\hline
\textcolor{darkblue}{\textbf{Dataset}}: \drop{} \\
\textcolor{darkblue}{\textbf{Question}}: Which player scored the first points of the game? \\
\textcolor{darkblue}{\textbf{Passage:}} The Patriots clinched their fourth straight AFC East title with a close road win . After a scoreless first quarter , the Jaguars responded to a \hl{Gostkowski} field goal with a Maurice Jones - Drew touchdown run . The Patriots challenged the play , as Jones - Drew appeared to fall down at the line of scrimmage , but the ruling on the field was upheld . New England came back before the halftime to retake the lead at 10 - 7 on a Dillon one - yard touchdown run . The Patriots maintained their lead as the teams traded touchdowns in the second half , including another touchdown by Jones - Drew . A David Garrard fumble with 1:55 left in the fourth quarter , recovered by safety Rodney Harrison , sealed the Patriots ' 11th win of the season . \\
\textcolor{darkblue}{\textbf{\textsc{gt:}}} [`Gostkowski'] \\
\textcolor{darkblue}{\textbf{\erm{} Answer:}} Maurice Jones \\
\textcolor{darkblue}{\textbf{\kd{}-Aug Answer:}} Gostkowski \\
\hline
\end{tabular}
\caption{Examples of test cases where \kd{}-based methods improve over plain \erm{}.}
\label{table:qualitative}
\end{table*}

\begin{table*}[h]
\small
\centering
\begin{tabular}{l|ccc}
\multicolumn{1}{c|}{\textbf{Method}} & \textbf{Source Datasets} & \textbf{Learning Rate} & \textbf{$\#$ of Epochs} \\
\Xhline{2\arrayrulewidth}
\multirow{6}{*}{\erm{}} & \textsc{c, d, l, p, v} & 3e-5 & 2 \\
& \textsc{d, l, p, v, w} & 1e-5 & 2 \\
& \textsc{l, p, v, w, c} & 1e-5 & 2 \\
& \textsc{p, v, w, c, d} & 7e-6 & 2 \\
& \textsc{v, w, c, d, l} & 1e-5 & 2 \\
& \textsc{w, c, d, l, p} & 1e-5 & 2 \\
\hline
\multirow{6}{*}{Domain-Adv} & \textsc{c, d, l, p, v} & 5e-5 & 2 \\
& \textsc{d, l, p, v, w} & 7e-5 & 2 \\
& \textsc{l, p, v, w, c} & 7e-5 & 2 \\
& \textsc{p, v, w, c, d} & 7e-5 & 2 \\
& \textsc{v, w, c, d, l} & 9e-5 & 2 \\
& \textsc{w, c, d, l, p} & 9e-5 & 2 \\
\hline
\multirow{6}{*}{Episodic} & \textsc{c, d, l, p, v} & 9e-6 & 2 \\
& \textsc{d, l, p, v, w} & 9e-6 & 2 \\
& \textsc{l, p, v, w, c} & 1e-5 & 2 \\
& \textsc{p, v, w, c, d} & 7e-6 & 2 \\
& \textsc{v, w, c, d, l} & 7e-6 & 2 \\
& \textsc{w, c, d, l, p} & 9e-6 & 2 \\
\hline
\multirow{6}{*}{\mldg{}} & \textsc{c, d, l, p, v} & 9e-6 & 2 \\
& \textsc{d, l, p, v, w} & 1e-5 & 2 \\
& \textsc{l, p, v, w, c} & 1e-5 & 2 \\
& \textsc{p, v, w, c, d} & 3e-5 & 1 \\
& \textsc{v, w, c, d, l} & 9e-6 & 2 \\
& \textsc{w, c, d, l, p} & 9e-6 & 2 \\
\hline
\multirow{6}{*}{\kd{} (gold-only)} & \textsc{c, d, l, p, v} & 7e-5 & 2 \\
& \textsc{d, l, p, v, w} & 3e-5 & 2 \\
& \textsc{l, p, v, w, c} & 3e-5 & 2 \\
& \textsc{p, v, w, c, d} & 5e-5 & 2 \\
& \textsc{v, w, c, d, l} & 5e-5 & 2 \\
& \textsc{w, c, d, l, p} & 3e-5 & 2 \\
\hline
\multirow{6}{*}{\kd{} (augmented)} & \textsc{c, d, l, p, v} & 5e-5 & \textit{synthetic:} 1, \textit{gold:} 2 \\
& \textsc{d, l, p, v, w} & 5e-5 & \textit{synthetic:} 1, \textit{gold:} 2 \\
& \textsc{l, p, v, w, c} & 3e-5 & \textit{synthetic:} 1, \textit{gold:} 1 \\
& \textsc{p, v, w, c, d} & 5e-5 & \textit{synthetic:} 1, \textit{gold:} 2 \\
& \textsc{v, w, c, d, l} & 5e-5 & \textit{synthetic:} 1, \textit{gold:} 2 \\
& \textsc{w, c, d, l, p} & 9e-6 & \textit{synthetic:} 1, \textit{gold:} 2 \\
\hline
\multirow{6}{*}{\kd{} (gold) w/ Domain-Adv} & \textsc{c, d, l, p, v} & 7e-5 & 2 \\
& \textsc{d, l, p, v, w} & 9e-5 & 2 \\
& \textsc{l, p, v, w, c} & 9e-5 & 2 \\
& \textsc{p, v, w, c, d} & 7e-5 & 2 \\
& \textsc{v, w, c, d, l} & 1e-4 & 2 \\
& \textsc{w, c, d, l, p} & 1e-4 & 2 \\
\hline
\multirow{6}{*}{\kd{} (gold) w/ Episodic} & \textsc{c, d, l, p, v} & 3e-5 & 2 \\
& \textsc{d, l, p, v, w} & 3e-5 & 2 \\
& \textsc{l, p, v, w, c} & 3e-5 & 2 \\
& \textsc{p, v, w, c, d} & 3e-5 & 2 \\
& \textsc{v, w, c, d, l} & 3e-5 & 2 \\
& \textsc{w, c, d, l, p} & 3e-5 & 2 \\
\hline
\multirow{6}{*}{\kd{} (gold) w/ \mldg{}} & \textsc{c, d, l, p, v} & 5e-5 & 2 \\
& \textsc{d, l, p, v, w} & 5e-5 & 2 \\
& \textsc{l, p, v, w, c} & 5e-5 & 2 \\
& \textsc{p, v, w, c, d} & 5e-5 & 2 \\
& \textsc{v, w, c, d, l} & 7e-5 & 2 \\
& \textsc{w, c, d, l, p} & 5e-5 & 2 \\
\end{tabular}
\caption{Optimal values of shared hyperparameters (learning rate, $\#$ of epochs). {\sc c}:~\searchqa{}, {\sc d}:~\squad{}, {\sc l}:~{\sc n}atural{\sc q}uestions (\nq{}), {\sc p}:~\hotpotqa{}, {\sc v}:~\triviaqa{}, {\sc w}:~\newsqa{}.}
\label{table:common-hyperparams}
\end{table*}

\subsection{Further Commentary on \kd{} Results}
\label{appendix:additional-kd-results}
The \bert{}-\textit{base} model fine-tuned using \erm{} has an F1 score of 75.0 on the in-domain dev set; the corresponding number for the \bert{}-\textit{large} teacher is 77.3.
As Table~\ref{table:id_vs_ood} shows, the augmented \kd{} student has an F1 score of 77.2, indicating that additional synthetic questions facilitated almost perfect distillation from \textit{large} to \textit{base} on our source domains.

\subsection{Model Selection}
\label{appendix:model-selection}
We use a training batch size of 32 for all models (with gradient accumulation when necessary).
To ensure fair comparison, we train and validate all methods on a large set of learning rates: $\{1,3,5,7,9\}\times10^{-6}$, $\{1,3,5,7,9\}\times10^{-5}$ and $\{1,3,5\}\times10^{-4}$, as the optimal learning rate varied drastically across different methods in our validation experiments. All models are trained for two epochs; we select the best of the epoch 1 and the epoch 2 checkpoint on the validation set for final evaluation on the \ood{} test set.
In Table~\ref{table:common-hyperparams} we provide the optimal values of these two hyperparameters for all models.

Table~\ref{table:method-specific-hyperparams} provides the optimal combinations of method-specific hyperparameters.
Below is a brief description of each:
\begin{enumerate}
    \item $\lambda_{adv}$ (Domain-Adv): This is the weight of the adversarial loss of the domain classifier. The main \erm{} loss has a fixed weight of 1.
    \item $\lambda_{erm}$, $\lambda_{episodic}$ (Episodic): The relative weights of the \erm{} loss and the episodic training loss in their convex combination.
    \item $\beta$ (\mldg{}): The weight of the meta-test loss during meta-optimization (second-order differentiation). The meta-train loss has a fixed weight of 1. 
    \item $\tau$ (\kd{}): Temperature (Eq.~\ref{equation:kd}).
\end{enumerate}

\subsection{Infrastructure and Computation}
\label{appendix:infrastructure}
We run all experiments on a single V100 GPU with 32GB memory.
A vast majority of the runs take less than a day; the longest ones take less than 48h. 

\begin{table*}[h]
\small
\centering
\begin{tabular}{l|ccc}
\multicolumn{1}{c|}{\textbf{Method}} & \textbf{Hyperparameters} & \textbf{Grid} & \textbf{Optimal} \\
\Xhline{2\arrayrulewidth}
Domain-Adv & $\lambda_{adv}$ & $\{0.01, 0.1, 1.0\}$ & $0.1$ \\
\hline
\multirow{2}{*}{Episodic} & \multirow{2}{*}{$(\lambda_{erm}, \lambda_{episodic})$} & $\lambda_{erm} \in \{0.25, 0.5, 0.75, 0.9\}$ & \multirow{2}{*}{$(0.75, 0.25)$} \\
 & & $\lambda_{episodic} = 1 - \lambda_{erm}$ & \\
\hline
\mldg{} & $\beta$ & \{1\} & 1 \\
\hline
\kd{} (gold) & $\tau$ & $\{1, 2, 4\}$ & $2$ \\
\hline
\kd{} (augmented) & $\tau$ & $\{4\}$ & $4$ \\
\hline
\kd{} (gold) w/ Domain-Adv & ($\tau, \lambda_{adv}$) & $\{1, 2, 4\} \times \{0.01, 0.1, 1.0\}$ & ($1, 0.01$) \\
\hline
\multirow{2}{*}{\kd{} (gold) w/ Episodic} & \multirow{2}{*}{($\tau, \lambda_{erm}, \lambda_{episodic}$)} & $\{1, 2, 4\} \times \lambda_{erm} \in \{0.25, 0.5, 0.75, 0.9\}$ & \multirow{2}{*}{($1, 0.75, 0.25$)} \\
 & & $\lambda_{episodic} = 1 - \lambda_{erm}$ & \\
\hline
\kd{} (gold) w/ \mldg{} & ($\tau, \beta$) & $\{1, 2, 4\} \times \{1\}$ & ($4, 1$) \\
\hline
\end{tabular}
\caption{Optimal values of hyperparameters specific to different training methods and the respective search grids.}
\label{table:method-specific-hyperparams}
\end{table*}

\end{document}